\documentclass[10pt,twocolumn,letterpaper]{article}

\usepackage{iccv}
\usepackage{times}
\usepackage{epsfig}
\usepackage{graphicx}
\usepackage{amsmath}
\usepackage{amssymb}

\usepackage{algorithm}
\usepackage{algpseudocode}

\usepackage{multirow}
\usepackage{comment}
\usepackage{xcolor}
\usepackage{pifont}

\usepackage{amsthm}
\newtheorem{theorem}{Theorem}
\newtheorem{lemma}[theorem]{Lemma}
\newcommand{\figref}[1]{Figure~\ref{#1}}
\newcommand{\tabref}[1]{Table~\ref{#1}}
\newcommand{\algoref}[1]{Algorithm~\ref{#1}}
\newcommand{\thmref}[1]{Theorem~\ref{#1}}
\newcommand{\secref}[1]{Section~\ref{#1}}

\definecolor{pink}{RGB}{255, 192, 201}
\definecolor{lightgreen}{RGB}{183, 215, 168}
\definecolor{skyblue}{RGB}{163, 194, 244}

\usepackage[pagebackref=true,breaklinks=true,letterpaper=true,colorlinks,bookmarks=false]{hyperref}

\iccvfinalcopy 

\def\iccvPaperID{1803} 

\ificcvfinal\fi

\begin{document}

\title{Self-Supervised Adaptation for Video Super-Resolution}

\author{Jinsu Yoo and Tae Hyun Kim\\
Hanyang University, Seoul, Korea\\
{\tt\small \{jinsuyoo,taehyunkim\}@hanyang.ac.kr}
}

\maketitle
\ificcvfinal\thispagestyle{empty}\fi

\setcounter{footnote}{-1}

\begin{abstract}

Recent single-image super-resolution (SISR) networks, which can adapt their network parameters to specific input images, have shown promising results by exploiting the information available within the input data as well as large external datasets.
However, the extension of these self-supervised SISR approaches to video handling has yet to be studied. 
Thus, we present a new learning algorithm that allows conventional video super-resolution (VSR) networks to adapt their parameters to test video frames without using the ground-truth datasets.
By utilizing many self-similar patches across space and time,
we improve the performance of fully pre-trained VSR networks and produce temporally consistent video frames.
Moreover, we present a test-time knowledge distillation technique that accelerates the adaptation speed with less hardware resources.
In our experiments, we demonstrate that our novel learning algorithm can fine-tune state-of-the-art VSR networks and substantially elevate performance on numerous benchmark datasets.

\end{abstract}



\section{Introduction}

Super-resolution (SR) aims to recover high-resolution (HR) images or videos given their low-resolution (LR) counterparts. Moreover, many techniques have been widely used in various areas, including medical imaging, satellite imaging, and electronics (\eg, smartphones, TV).
However, recovering high-quality HR images from LR images is ill-posed and challenging.
To solve this problem, early researchers have investigated reconstruction-based \cite{marquina2008image, sun2008image} and exemplar-based \cite{chang2004super, freeman2002example} methods. Dong \etal \cite{dong2015srcnn} proposed the use of convolutional neural networks (CNNs) to solve single-image super-resolution (SISR) for the first time. Kappeler \etal \cite{kappeler2016vsrnet} extended this neural approach to the video super-resolution (VSR) task. Since then, many deep learning-based approaches have been introduced~\cite{kim2016vdsr, kim2016drcn, ledig2017srgan, tian2020tdan, wang2019edvr, zhang2018rcan}. To benefit from the generalization ability of deep learning, most of these SR networks are trained with large external datasets.
Meanwhile, some researchers have studied ``zero-shot" SR approaches, which exploit similar patches across different scales within the input image and video~\cite{glasner2009sisr, huang2015selfex, shocher2018zssr, shahar2011space}. However, searching for LR-HR pairs of patches within LR images is also difficult. Moreover, the number of self-similar patches critically decreases as the scale factor increases~\cite{zontak2011internal}, and the searching problem becomes more challenging. To improve the performance by easing the searching phase, a coarse-to-fine approach is widely used~\cite{huang2015selfex, shocher2018zssr}.
Recently, several neural approaches that utilize external and internal datasets have been introduced and have produced satisfactory results~\cite{park2020mlsr, soh2020mzsr, lee2021dynavsr}. However, these methods remain bounded to the smaller scaling factor possibly because of a conventional approach to self-supervised data pair generation during the test phase. 

Therefore, we aim to develop a new learning algorithm that allows to explore the information available within given input video frames
without using clean ground-truth frames at test time.
Specifically, we utilize the space-time patch-recurrence over consecutive video frames and adapt the network parameters of a pre-trained VSR network for the test video during the test phase. 

To train the network without relying on ground-truth datasets, we present a new dataset acquisition technique for self-supervised adaptation.
Conventional self-supervised approaches are limited to handling a relatively small scaling factor (\eg, $\times$2), whereas our proposed technique allows a large upscaling factor (\eg, $\times$4).
Specifically, we utilize initially restored video frames from the fully pre-trained VSR networks to generate training targets for the test-time adaptation. In this manner, we can naturally combine external and internal data-based methods and elevate the performance of the pre-trained VSR networks. 
We summarize our contributions as follows:
\begin{itemize} 
    \item We propose a self-supervised adaptation algorithm that can exploit the internal statistics of input videos, and provide theoretical analysis.
    \item Our pseudo datasets allow a large scaling factor for the VSR task without gradual manner.
    \item We introduce a simple yet efficient test-time knowledge distillation strategy.
    \item We conduct extensive experiments with state-of-the-art VSR networks and achieve consistent improvement on public benchmark datasets by a large margin.
\end{itemize}

\section{Related works}

\paragraph{External-data-based SR.}

As three-layered deep CNNs are known to perform better than traditional model-based methods for the SISR task, several deep learning-based SR networks have been proposed. Kim \etal introduced a deeper network with residual learning \cite{kim2016vdsr} and also proposed an efficient learning scheme with the recursive parameter reuse technique \cite{kim2016drcn}. By removing unnecessary modules, such as batch normalization, and by stabilizing the learning procedure with residual learning, Lim \etal \cite{lim2017edsr} presented an even deeper network. Zhang \etal \cite{zhang2018rcan} brought channel attention to the network to make feature learning concise and proposed a residual-in-residual concept for stable learning. Recently, Dai \etal \cite{dai2019san} further enhanced the attention module with second-order statistics.

Starting from the neural approach for VSR tasks by Kappeler \etal \cite{kappeler2016vsrnet}, researchers have focused on utilizing redundant information among neighboring frames. 
To do so, Sajjadi \etal \cite{sajjadi2018frvsr} proposed a frame-recurrent network by adding a flow estimation network to convey temporal information. 
Instead of adding a motion compensation module, Jo \etal \cite{jo2018duf} directly upscaled input videos by using estimated dynamic filters. 
Xue \etal \cite{xue2019toflow} trained the flow estimation module to make it task oriented by jointly training the flow estimation and image-enhancing networks for various video restoration tasks (\eg, denoising and VSR). 
Instead of stacking neighboring frames, Haris \etal \cite{haris2019rbpn} proposed an iterative restoration framework by using recurrent back-projection architecture.
Tian \etal \cite{tian2020tdan} devised a deformable layer to align frames in a feature space as an alternative to optical flow estimation; this layer is further enhanced in the work of Wang \etal~\cite{wang2019edvr}. 

Although the existing methods have considerably improved network performance through training with large external datasets, they have limited capacity to exploit useful information within test input data. 
Our proposed method is embedded on top of pre-trained networks in a supervised manner to maximize the generalization ability of deep networks and also utilize the internal information within input test videos. 

\paragraph{Internal-data-based SR.}

Among pioneering works on the internal data-based SR, Glasner \etal~ \cite{glasner2009sisr} generated HR images solely from a single LR image by utilizing recurring patches within same and across scales.
Zontak \etal~\cite{zontak2011internal} deeply analyzed the patch-recurrence property within a single image. 
Huang \etal \cite{huang2015selfex} further handled geometrically transformed similar patches to enlarge the searching space of patch-recurrence.
Shahar \etal \cite{shahar2011space} extended the internal SISR method to the VSR task by observing that similar patches tend to repeat across space and time among neighboring video frames. 

Recently, Shocher \etal \cite{shocher2018zssr} trained an SR network given a test input LR image by using the internal data statistics for the first time. 
To solve ``zero-shot" video frame interpolation (temporal SR), Zuckerman \etal \cite{zuckerman2020vtzsr} exploited patch-recurrence not only within a single image but also across the temporal space.

More recently, Park \etal \cite{park2020mlsr} and Soh \etal \cite{soh2020mzsr} exploited the advantages of external and internal datasets by using meta-learning and Lee \etal \cite{lee2021dynavsr} further applied the technique to the VSR task.
Through meta-training, network parameters can be adapted to the given test image quickly, and the proposed methods can shorten the self-supervised learning procedure.
However, these methods are limited to the small scaling factor (\eg, $\times$2) because their conventional pseudo datasets generation strategy lacks high-frequency details to be exploited.

In contrast to existing studies, we aim to adapt the parameters of pre-trained VSR networks with a given LR video sequence at test time for a larger scaling factor without a coarse-to-fine manner, and we introduce a new strategy to generate pseudo datasets for self-supervised learning.

\paragraph{Knowledge distillation.}

Knowledge distillation from a bigger (teacher) network to a smaller (student) one was first suggested by Hinton \etal \cite{hinton2015distilling}.
They trained a shallow network to imitate a deeper network for the classification task while keeping high performance; many follow-up studies have been introduced~\cite{romero2014fitnets, mirzadeh2020improved, zhang2018dml, huang2017like}.

Recently, a few researchers have attempted to use knowledge distillation techniques for the SR task.
He \etal \cite{he2020fakd} proposed affinity-based distillation loss to bound space of the features; this approach enables further suitable loss to the regression task.
Lee \etal \cite{lee2020learning} constructed the teacher architecture with auto-encoder 
and trained the student network to resemble the decoder part of the teacher.

In contrast to conventional approaches that use the notion of knowledge distillation during the training phase, we distill the knowledge of a bigger network during test time to a smaller network to boost the adaptation efficiency.

\section{Proposed method}

In this section, we present a self-supervised learning approach based on the patch-recurrence property and provide theoretical analysis on the proposed algorithm.

\subsection{Patch-recurrence among video frames}
\label{3_1}

\begin{figure}[t]
    \centering
    \includegraphics[width=1\linewidth]{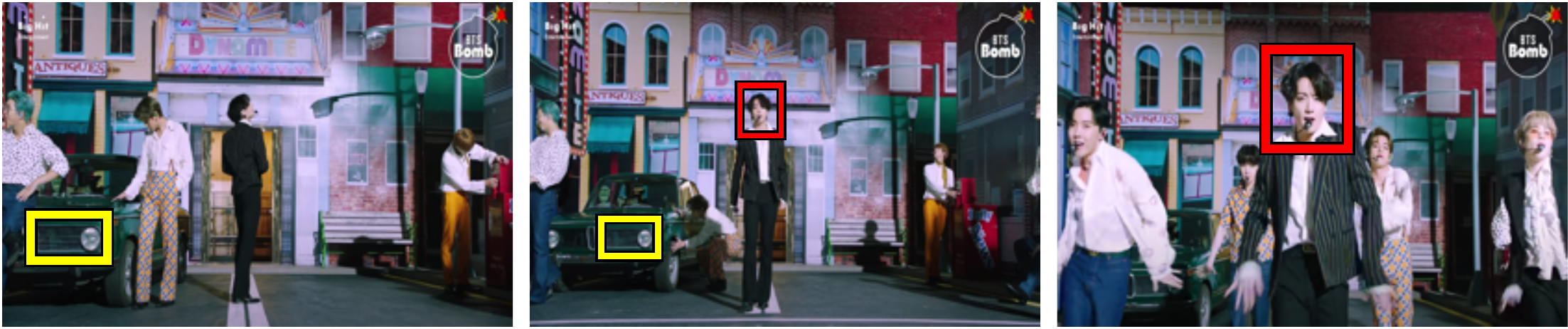}
    \caption[]{Recurring patches in a real video.\footnotemark Many similar patches of different scales can be observed across multiple consecutive video frames by the camera motion (yellow box) and moving objects (red box).}
    \label{fig_bts}
\end{figure}

Many similar patches 
exist within a single image~\cite{glasner2009sisr, zontak2011internal}.
Moreover, the number of these self-similar patches increases when we deal with a video rather than a single image~\cite{shahar2011space}.
As shown in \figref{fig_bts}, forward and backward motions of camera and/or objects generate recurring patches of different scales across multiple frames, which are crucial for the SR task. Specifically, larger patches include more detailed information than the corresponding smaller ones among neighboring frames, and these additional details facilitate the enhancement of the quality of the smaller ones, as introduced in~\cite{shocher2018zssr,park2020mlsr}.

\footnotetext{\href{https://www.youtube.com/watch?v=Bc3OibcQNHA}{Dynamite - BTS}}

\subsection{Pseudo datasets for large-scale VSR}
\label{3_2}

\begin{figure}[t]
    \centering
    \includegraphics[width=1\linewidth]{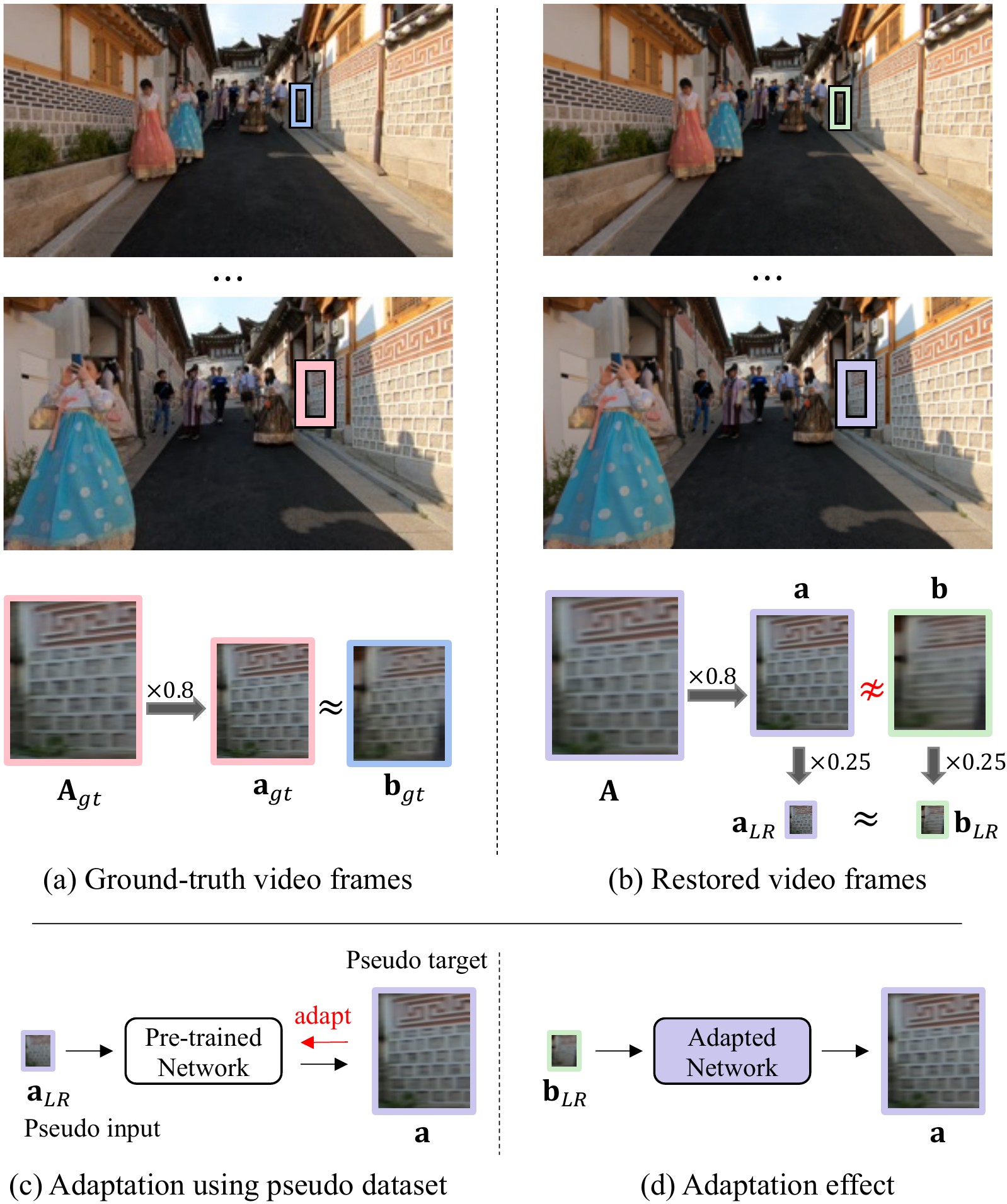}
    \caption{Our key observation. (a) - (b): Ground-truth and restored HR frames by EDVR~\cite{wang2019edvr} show patch-recurrence across difference scales. Unlike ground-truth frames, downscaled version of a large patch includes more details in the restored HR frames. (c) - (d): Our goal is increasing the resolution of a small patch $\textbf{b}_{LR}$ using the downscaled patch $\textbf{a}$ within the HR frames by adaptation.}
    \label{fig_corresponding}
\end{figure}

In exemplar-based SR, we need to search for corresponding large patches among neighboring frames to increase the resolution of a small patch by a large scaling factor.
For example, to increase the resolution of a $10 \times 10$ patch to a four-times enlarged one, we should find $40 \times 40$ patches within the LR inputs. However, these large target patches become scarce as the scaling factor increases~\cite{zontak2011internal}. 
Therefore, recent self-supervised approaches~\cite{shocher2018zssr,park2020mlsr,soh2020mzsr} are limited to a relatively small scaling factor (\eg, $\times$2).
One can take advantage of a gradual upscaling strategy, as suggested in \cite{huang2015selfex, shocher2018zssr}, but this coarse-to-fine approach greatly increases the adaptation time in the test stage.

To mitigate this problem and directly allow a large scaling factor,
we acquire pseudo datasets from the initially restored HR video frames by fully pre-trained VSR networks.

In \figref{fig_corresponding}, we illustrate how we organize datasets for the test-time adaptation without ground-truth targets. 
Our key observation in this work is that the visual quality of 
the downscaled version of a large patch and the corresponding small patch (\eg, $\textbf{a}_{gt}$ and $\textbf{b}_{gt}$ in \figref{fig_corresponding} (a)) is similar on the ground-truth video frames. However, this property does not hold with the HR frames predicted by conventional VSR networks, and the quality of the downscaled version of a large patch is much better than that of its corresponding small patch (\eg, $\textbf{a}$ and $\textbf{b}$ in \figref{fig_corresponding} (b)) because the LR version of the small patch ($\textbf{b}_{LR}$) includes minimal details and thus is non-discriminative for VSR networks to generate its high-quality counterpart. 
Furthermore, we discover that LR input of the small patch and a further downscaled version of the large patch become similar (\eg $\textbf{a}_{LR}$ and $\textbf{b}_{LR}$ in \figref{fig_corresponding} (b)) because the additional details in $\textbf{a}$ are also attenuated by the large downscaling to $\textbf{a}_{LR}$ from $\textbf{a}$.

Based on these findings,
we generate a new training dataset to improve the performance of the pre-trained network on the given input frames, and we use $\textbf{a}$ and $\textbf{a}_{LR}$ as our training target and input, respectively.
Using this dataset, we can fine-tune the pre-trained VSR networks, as shown in \figref{fig_corresponding} (c). Then, the fine-tuned network can increase the resolution of $\textbf{b}_{LR}$ with a corresponding HR patch $\textbf{a}$, thereby including additional details (\figref{fig_corresponding} (d)).
Note that, we generate the train set for the fine-tuning without using ground-truth frames; thus, our training targets become pseudo targets.
Moreover, given that our test-time adaptation method relies on pre-trained VSR networks on the large external datasets and initial restoration results with a large scaling factor, we can naturally combine internal and large external information and handle large scaling factors.

\subsection{Adaptation without patch-match}

In \figref{fig_corresponding}, we need to find a pair of corresponding patches (\ie $\textbf{A}$ and $\textbf{b}$) in the restored HR frames to enhance the quality of a patch $\textbf{b}$. However, finding these correspondences is a difficult task (\eg, optical flow estimation), which takes much time even with a naive patch-match algorithm \cite{barnes2010generalized}.

To alleviate this problem, we use a simple randomized scheme under the assumption that the distributions of $\textbf{a}_{LR}$ and $\textbf{b}_{LR}$ are similar, which improves $\textbf{b}$ without explicit searching for $\textbf{a}$.
Specifically, we randomly choose patch a $\textbf{A}$.
Then, we downscale $\textbf{A}$ to $\textbf{a}$, and $\textbf{a}$ to $\textbf{a}_{LR}$ in turn.
In this manner, we can generate a large number of pseudo train datasets. Statistically, patches with high patch-recurrence are likely to be included multiple times in our dataset.
Therefore, we can easily expose pairs of highly recurring patches across different scales to the VSR networks during adaptation, and the VSR networks can be fine-tuned without accurate correspondences if they are fully convolutional due to the translation equivariance property of CNNs~\cite{pmlr-v48-cohenc16}.

\if 
Let $y_m, y_n$ be the pseudo-targets for same content on different space-time location and $T(\cdot)$ represents the translation matrix from the location of $y_m$ to that of $y_n$'s. As we highlighted in \secref{3_2}, the pseudo-inputs $x_m$ and $x_n$ become non-discriminative ($x_m \approx x_n$) when the scaling factor is large enough (\eg, $\times 4$). As a result, the adaptation will minimize the loss $L$ as: $L(f_{\theta}(x_m), y_m) \approx L(f_{\theta}(x_n), y_{m}) \approx L(f_{\theta}(x_n), T(y_{m}))$. 
That is, similar pseudo-inputs share various pseudo-targets regardless of the space-time location, and the adapted network will give a single (similar) optimized output for the pseudo-inputs.
Notably, our algorithm has high applicability as any conventional CNNs-based deep learning-based network can be utilized.
\fi

\subsection{Overall flow}
\label{3_3}

\begin{algorithm}
\caption{Self-Supervised Adaptation}
\begin{algorithmic}[1]
    \Require Pre-trained VSR network $\textbf{f}_{\theta}$, test LR video frames $\{\textbf{X}_t\}$, initially restored frames $\{\textbf{Y}_t\}$
    
    \For{number of adaptations}
    
        \State $\text{Sample a frame: } \textbf{Y} \sim \{\textbf{Y}_t\}$ 
        
        \State $\text{Randomly crop patch } \textbf{Y}_p \text{ from } \textbf{Y}$ 
        
        \State $\text{Acquire pseudo target } \textbf{y} \text{ by randomly downscaling } \textbf{Y}_p$ 
        
        \State $\text{Acquire pseudo input } \textbf{y}_{LR} \text{ by downscaling } \textbf{y}$ 
        
        \State $\text{Compute gradient: } \nabla_{\theta}||\textbf{f}_{\theta}(\textbf{y}_{LR}) - \textbf{y}||_2^2$
        
        \State $\text{Update } \theta \text{ using the gradient-based learning rule}$
        
    \EndFor
    
    \State \Return $\{\textbf{f}_{\theta}(\textbf{X}_t)\}$
\end{algorithmic}
\label{algo_ssa}
\end{algorithm} 

The overall adaptation procedure of the proposed method is described in \algoref{algo_ssa}.
We first obtain the initial super-resolved frames $\{\textbf{Y}_t\}$ using a pre-trained VSR network $\textbf{f}_{\theta}$.
Next, we randomly select a frame $\textbf{Y}$ from the HR sequence $\{\textbf{Y}_t\}$, and crop a patch $\textbf{Y}_p$ from $\textbf{Y}$ randomly.
Then, the random patch $\textbf{Y}_p$ is downscaled by a random scaling factor to generate the pseudo target $\textbf{y}$. 
Thus, we can generate a corresponding pseudo LR input $\textbf{y}_{LR}$ by simply downscaling the pseudo target $\textbf{y}$ with the known desired scaling factor (\eg, $\times$4). 
By using this pseudo dataset, we update the network parameters by minimizing the distance between the pseudo target $\textbf{y}$ and the network output (\ie, $\textbf{f}_{\theta}(\textbf{y}_{LR})$) based on the mean squared error (MSE). The network can be optimized with a conventional gradient-based optimizer, such as Adam~\cite{kingma2014adam}, and we repeat these steps until convergence.
Finally, we can render the enhanced outputs by using the updated network parameters.

\subsection{Theoretical analysis}
\label{3_4}

In this section, we analyze the adaptation procedure to understand the principle of the proposed method in~\algoref{algo_ssa}.

\subsubsection*{Adaptation performance}

\begin{figure}[]
    \centering
    \includegraphics[width=\linewidth]{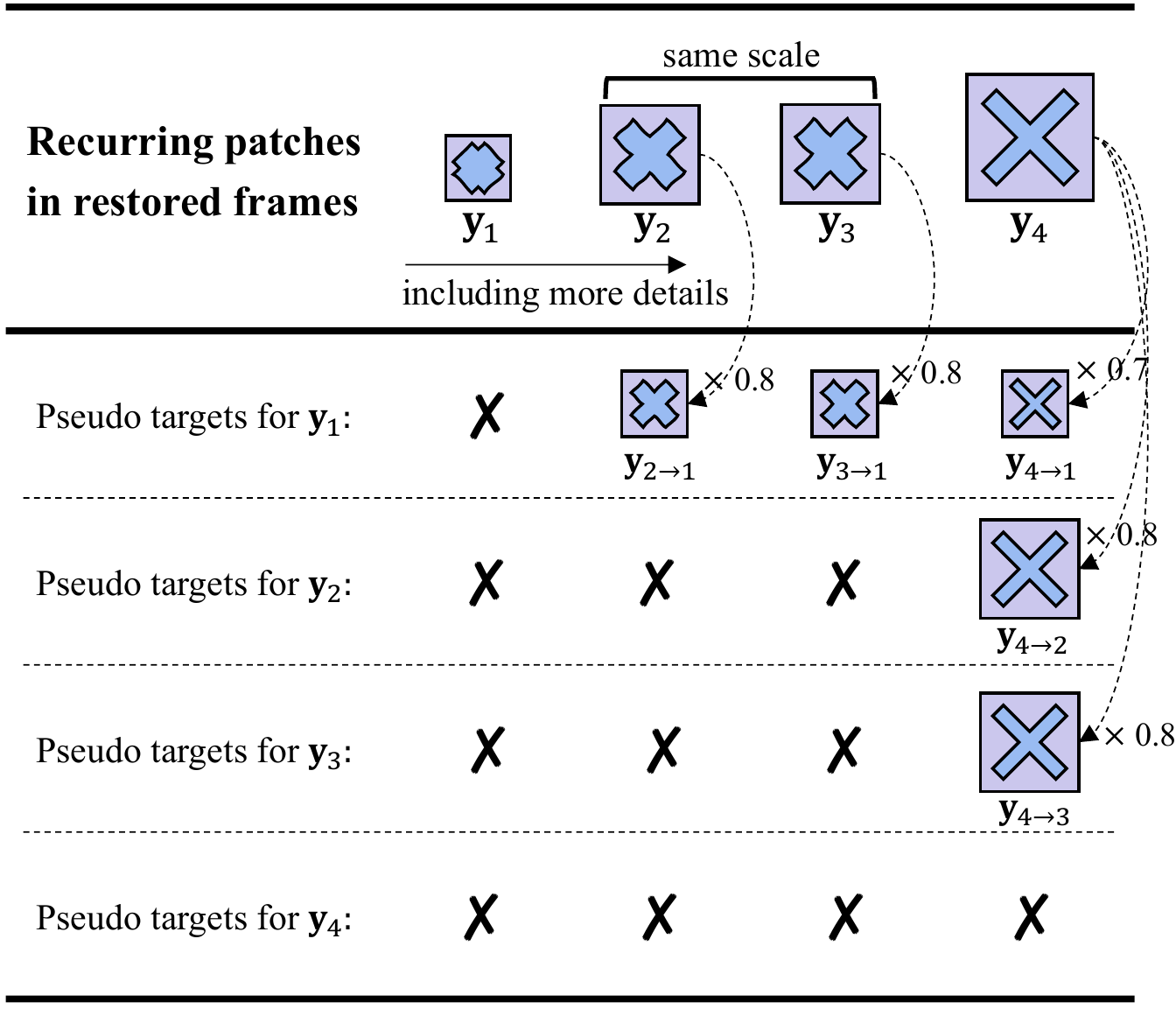}
    \caption{Illustration of pseudo target generation. Our adaptation procedure can selectively utilize patches with more details by downscaling bigger patches.}
    \label{fig_analyze}
\end{figure}

In~\figref{fig_corresponding}, we observed that larger patches can help improve the quality of the corresponding but smaller ones. We analyzed this observation more concretely.
Assume that we have $k$ similar restored HR patches from various scales, and they are sorted from smallest to largest ones as illustrated in the top of \figref{fig_analyze}.
Then, we can guarantee that the quality of the SR results by our adaptation algorithm is better than the initially restored results from the pre-trained baseline.
\begin{theorem}
\label{thm1}
The restoration quality of recurring patches improves after the adaptation.
\end{theorem}

\begin{proof}
As we assume that corresponding HR patches are sorted, larger versions of a patch $\textbf{y}_{m}$ are $\{\textbf{y}_{i}\}_{i=m+1}^{k}$ when $1 \leq m < k$.
Using known SR kernel (\eg, bicubic), we can easily downscale these $(k-m)$ larger patches (\ie $\{\textbf{y}_{i}\}_{i=m+1}^{k}$) and generate $\{\textbf{y}_{i \to m}\}_{i=m+1}^{k}$ where the size of $\textbf{y}_{i \to m}$ equals that of $\textbf{y}_{m}$ (see \figref{fig_analyze}).

Note that, we acquire these pseudo targets $\{\textbf{y}_{i \to m}\}_{i=m+1}^{k}$ by using downscaling in \algoref{algo_ssa}, and thus the pseudo targets include more image details.
Accordingly, under an assumption that $\textbf{y}_m^{LR}$ and $\{\textbf{y}_{i \to m}\}^{LR}$ are identical, which are LR versions of $\textbf{y}_m$ and $\{\textbf{y}_{i \to m}\}$ by downscaling with the given large scaling factor (\eg, $\times$4), our \algoref{algo_ssa} will minimize the MSE loss for the patch $\textbf{y}_m$ as:
\begin{equation}
    \operatorname*{argmin}_{\theta} \frac{1}{k-m}\sum_{i=m+1}^{k} ||\textbf{f}_{\theta}(\textbf{y}_m^{LR}) - \textbf{y}_{i \rightarrow m}||_{2}^{2},
\end{equation}
where $\textbf{f}_\theta$ is the network to be adapted.
Then, we can update the parameter $\theta$, which results in
$\textbf{f}_\theta (\textbf{y}_m^{LR}) = \frac{1}{k-m}\sum_{i=m+1}^{k} \textbf{y}_{i \to m}$
for $m \in \{1, 2, ..., k-1\}$. Recall that patches $\textbf{y}_{i \to m}$ with larger $i$ includes more image details in our observation; thus, a newly restored version of $\textbf{f}_\theta (\textbf{y}_m^{LR})$ also includes more details than the initially restored patch $\textbf{y}_m$.
Meanwhile, the adaptation for $\textbf{y}_m$ naturally discards corresponding but smaller patches (\ie, $\{\textbf{y}_{i}\}_{i=1}^{m}$) because our proposed pseudo target generation is solely with the downscaling operation. 
\end{proof}


\subsubsection*{Space-time consistency}

Next, we provide analysis on the space-time consistency of the recurring patches. That is, we enforce consistency among recurring patches via our adaptation. 

\begin{lemma}
The adapted network generates consistent HR patches.
\end{lemma}

\begin{proof}
Assume that we have two corresponding patches $\textbf{y}_m$ and $\textbf{y}_n$ of the same size (\eg, $\textbf{y}_2$ and $\textbf{y}_3$ in \figref{fig_analyze}), then the adapted network parameter $\theta$ would predict the same results for these patches in accordance with \thmref{thm1} (\ie, $\frac{1}{k-n}\sum_{i=n+1}^{k} \textbf{y}_{i \to n}$ if $m < n$),
and the corresponding patches become identical.
\end{proof}

The above lemma shows that the corresponding HR patches by the adapted network are consistent.
This property is important because the adapted network is guaranteed to predict spatio-temporally consistent results.

\subsection{Efficient adaptation via knowledge distillation}
\label{3_5}

\begin{figure}[]
    \centering
    \includegraphics[width=\linewidth]{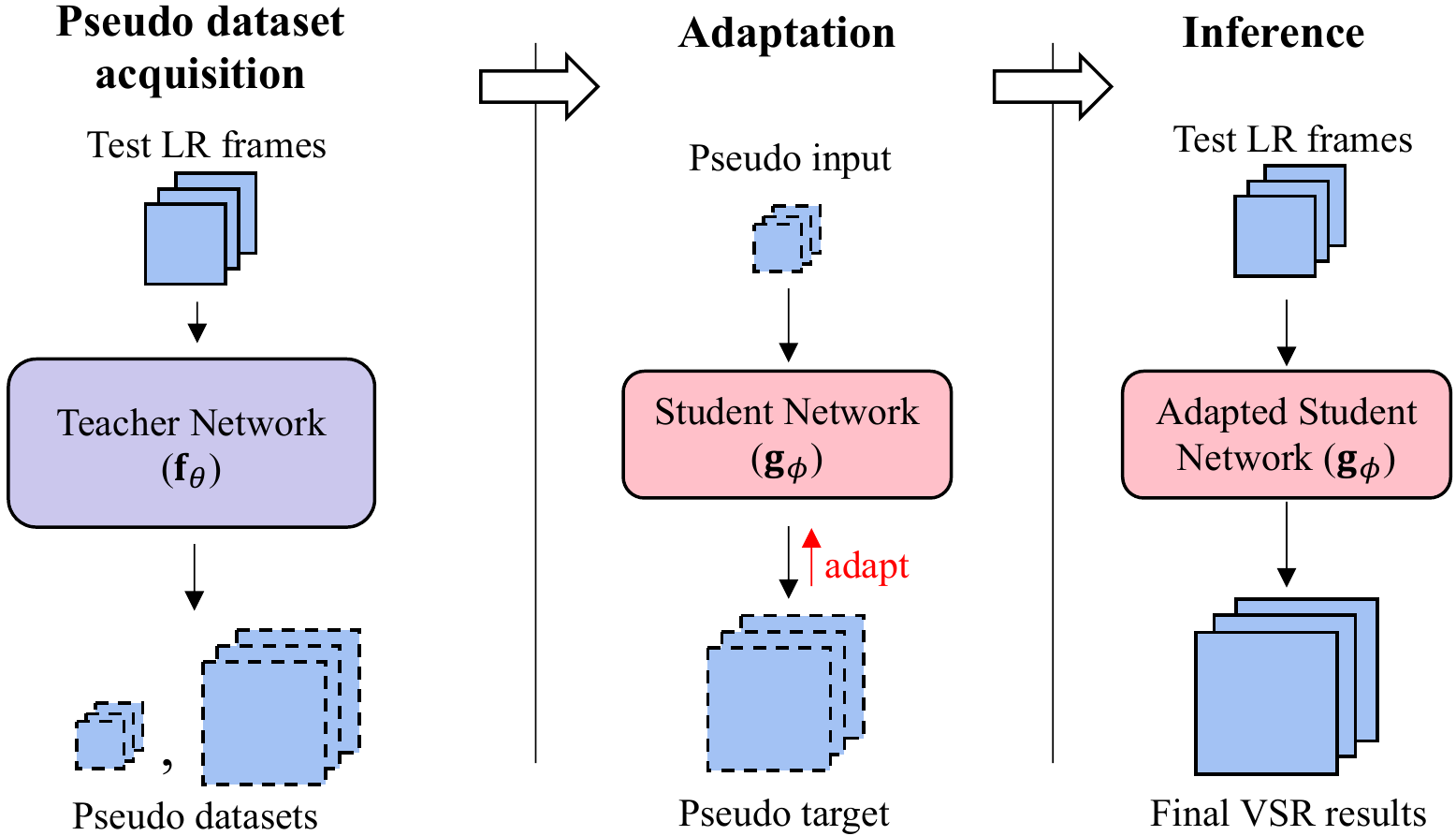}
    \caption{Illustration of the efficient adaptation using distillation.}
    \label{kd}
\end{figure}

\begin{algorithm}
\caption{Efficient Adaptation via Distillation}
\begin{algorithmic}[1]
    \Require Pre-trained teacher network $\textbf{f}_{\theta}$, pre-trained student network $\textbf{g}_{\phi}$, test LR video frames $\{\textbf{X}_t\}$, initially restored HR frames $\{\textbf{Y}_t\}$ by the teacher network 
    
    \For{number of adaptations}
    
        \State $\text{Sample a frame: }  \textbf{Y} \sim \{\textbf{Y}_t\}$    
        
        \State $\text{Randomly crop patch } \textbf{Y}_p \text{ from } \textbf{Y}$ 
        
        \State $\text{Acquire pseudo target } \textbf{y} \text{ by randomly downscaling } \textbf{Y}_p$ 
        
        \State $\text{Acquire pseudo input } \textbf{y}_{LR} \text{ by downscaling } \textbf{y}$ 
        
        \State $\text{Compute gradient: }\nabla_{\phi}||\textbf{g}_{\phi}(\textbf{y}_{LR}) - \textbf{y}||_{2}^{2}$
        
        \State $\text{Update } \phi \text{ using the gradient-based learning rule}$
    \EndFor
    
    \State \Return $\{\textbf{g}_{\phi}(\textbf{X}_t)\}$
\end{algorithmic}
\label{algo_fa}
\end{algorithm} 

Although our test-time adaptation algorithm in \algoref{algo_ssa} can elevate SR performance, it takes much time when the pre-trained network $\textbf{f}_\theta$ is large. To mitigate this problem, we introduce an efficient adaptation algorithm with the aid of a knowledge distillation technique as in \algoref{algo_fa}.
Specifically, we define teacher as a big network and student as a much smaller network (\figref{kd}). 
Conventional distillation~\cite{he2020fakd, lee2020learning} is performed during the training phase with ground-truth HR images, whereas we can distill useful information in test time solely with our generated pseudo datasets.
We find that our method without sophisticated techniques (\eg, feature distillation) reduces computational complexity whilst boosts the SR performance.

\section{Experiments}

In this section, we provide the quantitative and qualitative experimental results and demonstrate the performance of the proposed method. Please refer to our supplementary material for more results. Moreover, the code, dataset, and pre-trained models for the experiments are also included in the supplementary material.

\begin{table*}[]
\centering
\footnotesize
\begin{tabular}{lccccc}
\hline
\textbf{Method} & \textbf{\begin{tabular}[c]{@{}c@{}}Calendar\\ (PSNR/SSIM)\end{tabular}} & \textbf{\begin{tabular}[c]{@{}c@{}}City\\ (PSNR/SSIM)\end{tabular}} & \textbf{\begin{tabular}[c]{@{}c@{}}Foliage\\ (PSNR/SSIM)\end{tabular}} & \textbf{\begin{tabular}[c]{@{}c@{}}Walk\\ (PSNR/SSIM)\end{tabular}} & \textbf{\begin{tabular}[c]{@{}c@{}}Average\\ (PSNR/SSIM)\end{tabular}} \\ \hline
TOFlow \cite{xue2019toflow} & 22.44/0.7291 & 26.74/0.7367 & 25.24/0.7067 & 29.01/0.8776 & 25.86/0.7625 \\ \cline{2-6} 
TOFlow~\cite{xue2019toflow} + Adaptation & \begin{tabular}[c]{@{}c@{}}\textbf{22.73/0.7432}\end{tabular} & \begin{tabular}[c]{@{}c@{}}\textbf{26.97/0.7524}\end{tabular} & \begin{tabular}[c]{@{}c@{}}\textbf{25.43/0.7174}\end{tabular} & \begin{tabular}[c]{@{}c@{}}\textbf{29.20/0.8792}\end{tabular} & \begin{tabular}[c]{@{}c@{}}\textbf{26.08/0.7731}\end{tabular} \\ \hline
RBPN \cite{haris2019rbpn} & 23.95/0.8080 & 27.74/0.8057 & 26.22/0.7581 & 30.69/0.9112 & 27.15/0.8208 \\ \cline{2-6} 
RBPN \cite{haris2019rbpn} + Adaptation & \begin{tabular}[c]{@{}c@{}}\textbf{24.12/0.8139}\end{tabular} & \begin{tabular}[c]{@{}c@{}}\textbf{27.91/0.8133}\end{tabular} & \begin{tabular}[c]{@{}c@{}}\textbf{26.26/0.7591}\end{tabular} & \begin{tabular}[c]{@{}c@{}}\textbf{30.78/0.9113}\end{tabular} & \begin{tabular}[c]{@{}c@{}}\textbf{27.27/0.8244}\end{tabular} \\ \hline
EDVR \cite{wang2019edvr} & 24.05/0.8147 & 28.00/0.8122 & 26.34/0.7635 & 31.02/0.9152 & 27.35/0.8264 \\ \cline{2-6} 
EDVR \cite{wang2019edvr} + Adaptation & \begin{tabular}[c]{@{}c@{}}\textbf{24.37/0.8221}\end{tabular} & \begin{tabular}[c]{@{}c@{}}\textbf{28.13/0.8193}\end{tabular} & \begin{tabular}[c]{@{}c@{}}\textbf{26.39/0.7639}\end{tabular} & \begin{tabular}[c]{@{}c@{}}\textbf{31.22/0.9166}\end{tabular} & \begin{tabular}[c]{@{}c@{}}\textbf{27.53/0.8305}\end{tabular} \\ \hline
\end{tabular}

\begin{tabular}{lccccc}
\hline
\textbf{Method} & \textbf{\begin{tabular}[c]{@{}c@{}}Clip\_000\\ (PSNR/SSIM)\end{tabular}} & \textbf{\begin{tabular}[c]{@{}c@{}}Clip\_011\\ (PSNR/SSIM)\end{tabular}} & \textbf{\begin{tabular}[c]{@{}c@{}}Clip\_015\\ (PSNR/SSIM)\end{tabular}} & \textbf{\begin{tabular}[c]{@{}c@{}}Clip\_020\\ (PSNR/SSIM)\end{tabular}} & \textbf{\begin{tabular}[c]{@{}c@{}}Average\\ (PSNR/SSIM)\end{tabular}} \\ \hline
TOFlow \cite{xue2019toflow} & 27.83/0.7708 & 29.17/0.8107 & 32.00/0.8799 & 28.28/0.8157 & 29.32/0.8193 \\ \cline{2-6} 
TOFlow \cite{xue2019toflow} + Adaptation & \begin{tabular}[c]{@{}c@{}}\textbf{27.98/0.7772}\end{tabular} & \begin{tabular}[c]{@{}c@{}}\textbf{29.89/0.8280}\end{tabular} & \begin{tabular}[c]{@{}c@{}}\textbf{32.33/0.8870}\end{tabular} & \begin{tabular}[c]{@{}c@{}}\textbf{28.71/0.8289}\end{tabular} & \begin{tabular}[c]{@{}c@{}}\textbf{29.73/0.8303}\end{tabular} \\ \hline
RBPN \cite{haris2019rbpn} & 28.95/0.8226 & 31.47/0.8674 & 34.48/0.9225 & 30.02/0.8704 & 31.23/0.8707 \\ \cline{2-6} 
RBPN \cite{haris2019rbpn} + Adaptation & \begin{tabular}[c]{@{}c@{}}\textbf{29.04/0.8249}\end{tabular} & \begin{tabular}[c]{@{}c@{}}\textbf{31.78/0.8707}\end{tabular} & \begin{tabular}[c]{@{}c@{}}\textbf{34.72/0.9249}\end{tabular} & \begin{tabular}[c]{@{}c@{}}\textbf{30.10/0.8712}\end{tabular} & \begin{tabular}[c]{@{}c@{}}\textbf{31.41/0.8729}\end{tabular} \\ \hline
EDVR \cite{wang2019edvr} & 29.34/0.8374 & 33.55/0.9025 & 35.47/0.9341 & 31.45/0.9006 & 32.45/0.8937 \\ \cline{2-6} 
EDVR \cite{wang2019edvr} + Adaptation & \begin{tabular}[c]{@{}c@{}}\textbf{29.45/0.8386}\end{tabular} & \begin{tabular}[c]{@{}c@{}}\textbf{33.92/0.9052}\end{tabular} & \begin{tabular}[c]{@{}c@{}}\textbf{35.76/0.9369}\end{tabular} & \begin{tabular}[c]{@{}c@{}}\textbf{31.62/0.9021}\end{tabular} & \begin{tabular}[c]{@{}c@{}}\textbf{32.69/0.8957}\end{tabular} \\ \hline
\end{tabular}
\caption{Quantitative results of the proposed method using various baseline networks with $\times$4 upscaling factor on Vid4 (top) and REDS4 (bottom) dataset. The performance of the baseline networks is consistently boosted with our proposed adaptation.}
\label{tab_vid4_reds4}
\end{table*}

\subsection{Implementation details}
\label{4_1}

We implement our adaptation algorithm on the PyTorch framework and use NVIDIA GeForce RTX 2080Ti GPU for the experiments.

\paragraph{Baseline VSR networks and test dataset.} 

For our baseline VSR networks, we adopt three different VSR networks: TOFlow~\cite{xue2019toflow}, RBPN~\cite{haris2019rbpn}, and EDVR~\cite{wang2019edvr}.
Notably, EDVR is the state-of-the-art VSR approach at the time of submission. 
Each network is fully pre-trained with large external datasets, and 
we use publicly available pre-trained network parameters.\footnote{TOFlow and EDVR: \href{https://github.com/xinntao/BasicSR}{https://github.com/xinntao/BasicSR}, RBPN: \href{https://github.com/alterzero/RBPN-PyTorch}{https://github.com/alterzero/RBPN-PyTorch}}

To evaluate the performance of the proposed adaptation algorithm, 
we test our method on public test datasets, \ie, Vid4 and REDS4.
The Vid4 dataset~\cite{liu2011bayesian} includes four clips with 41, 34, 49, and 47 frames each. The video contains limited motion, and the ground-truth video still shows a certain amount of noise. 
The REDS4 test dataset~\cite{wang2019edvr} includes four clips from the original REDS dataset~\cite{nah2019reds}. 
The REDS dataset comprises 720$\times$1280 HR videos from dynamic scenes. 
It also contains a larger motion than Vid4, and each clip contains 100 frames.
Note that, none of these test datasets are used for pre-training the baseline networks.

\paragraph{Adaptation setting and evaluation metrics.} 

We minimize MSE loss using the Adam~\cite{kingma2014adam} in \algoref{algo_ssa} and \algoref{algo_fa}. Refer to our supplementary material and codes for detailed settings, including patch size, batch size, and learning rate. For each pseudo dataset generation procedure, we randomly choose a downscaling factor from 0.8 to 0.95. The number of adaptation iterations for the Vid4 and REDS4 datasets are 1K and 3K, respectively. All the experiments are conducted with a fixed upscaling factor (\ie, $\times$4), which is the most challenging setting in conventional VSR works. 


We evaluate the SR results in terms of peak signal-to-noise ratio (PSNR) and structure similarity (SSIM). In calculating the values, we convert the RGB channel into the YCbCr channel and use only the Y channel as suggested in~\cite{wang2019edvr}.
Moreover, to evaluate the temporal consistency of the restored frames, we use a pixel-wise error of the estimated optical flow (tOF) as introduced in~\cite{chu2020tecogan}.

\begin{figure*}[]
\centering
\includegraphics[width=\linewidth]{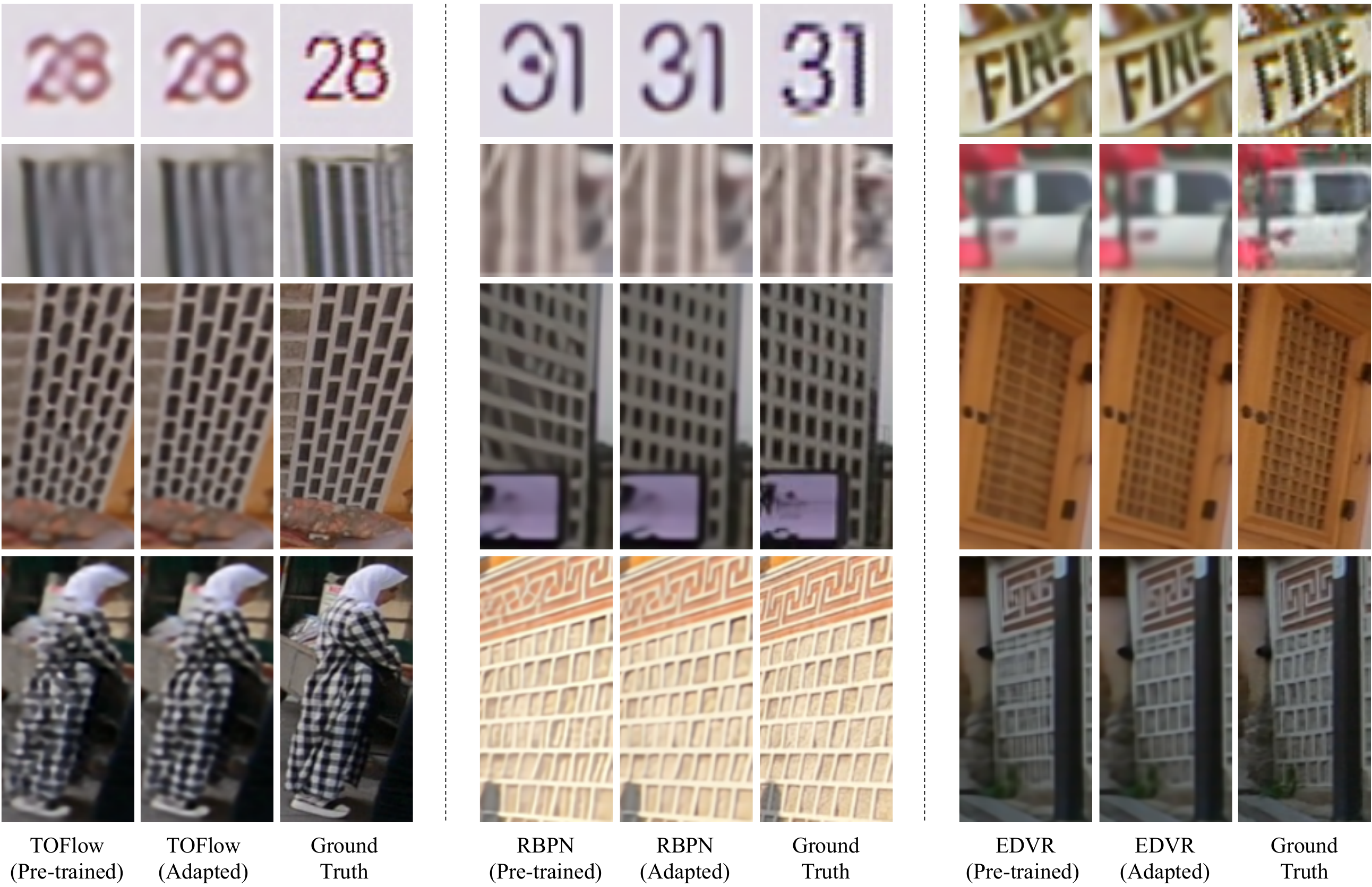}
\caption{Visual comparison. Initial VSR results with $\times$4 scaling factor by baseline networks (TOFlow, RBPN, and EDVR) and the adapted VSR results are compared on the Vid4 and REDS4 datasets. Our adaptation successfully restores the degraded parts from the baseline networks using recurring patches.}
\label{fig_visual}
\end{figure*}

\subsection{Restoration results} 

\paragraph{Quantitative and qualitative VSR results.}

In \tabref{tab_vid4_reds4}, we compare the SR performance before and after adaptation. TOFlow~\cite{xue2019toflow}, RBPN~\cite{haris2019rbpn}, and EDVR~\cite{wang2019edvr} are used as our baselines and evaluated on the Vid4 and REDS4 datasets. 
The proposed method consistently improves the SR performance over the baseline networks.
In particular, we observe a large margin on the REDS4 dataset because the REDS4 dataset includes more recurring patches (more frames and forward/backward motions than the Vid4 dataset). 

In \figref{fig_visual}, we provide visual comparisons. We see that the restored frames after using our adaptation algorithm show much clearer and sharper results than the initial results by the pre-trained baseline networks.
In particular, broken and distorted edges are well restored.

\paragraph{Temporal consistency.}

\begingroup
\setlength{\tabcolsep}{3.5pt}
\begin{table}[]
\centering
\footnotesize
\begin{tabular}{lcccc}
\hline
\textbf{Dataset} & \textbf{Adapting} & \textbf{\begin{tabular}[c]{@{}c@{}}TOFlow~\cite{xue2019toflow}\\ (tOF)\end{tabular}} & \textbf{\begin{tabular}[c]{@{}c@{}}RBPN~\cite{haris2019rbpn}\\ (tOF)\end{tabular}} & \textbf{\begin{tabular}[c]{@{}c@{}}EDVR~\cite{wang2019edvr}\\ (tOF)\end{tabular}} \\ \hline
\multirow{2}{*}{Vid4} & X & 0.2116 & 0.1470 & 0.1362 \\
 & O & \textbf{0.1954} & \textbf{0.1376} & \textbf{0.1277} \\ \hline
\multirow{2}{*}{REDS4} & X & 1.7378 & 1.0561 & 0.8024 \\
 & O & \textbf{1.3957} & \textbf{0.9981} & \textbf{0.6356} \\ \hline
\end{tabular}
\caption{Evaluating temporal consistencies in terms of tOF \cite{chu2020tecogan} before and after adaptation. Our proposed method largely improves the temporal consistency than all baselines. Lower score indicates better performance.}
\label{tab_tof}
\end{table}
\endgroup

\begin{figure}[]
\centering
\includegraphics[width=\linewidth]{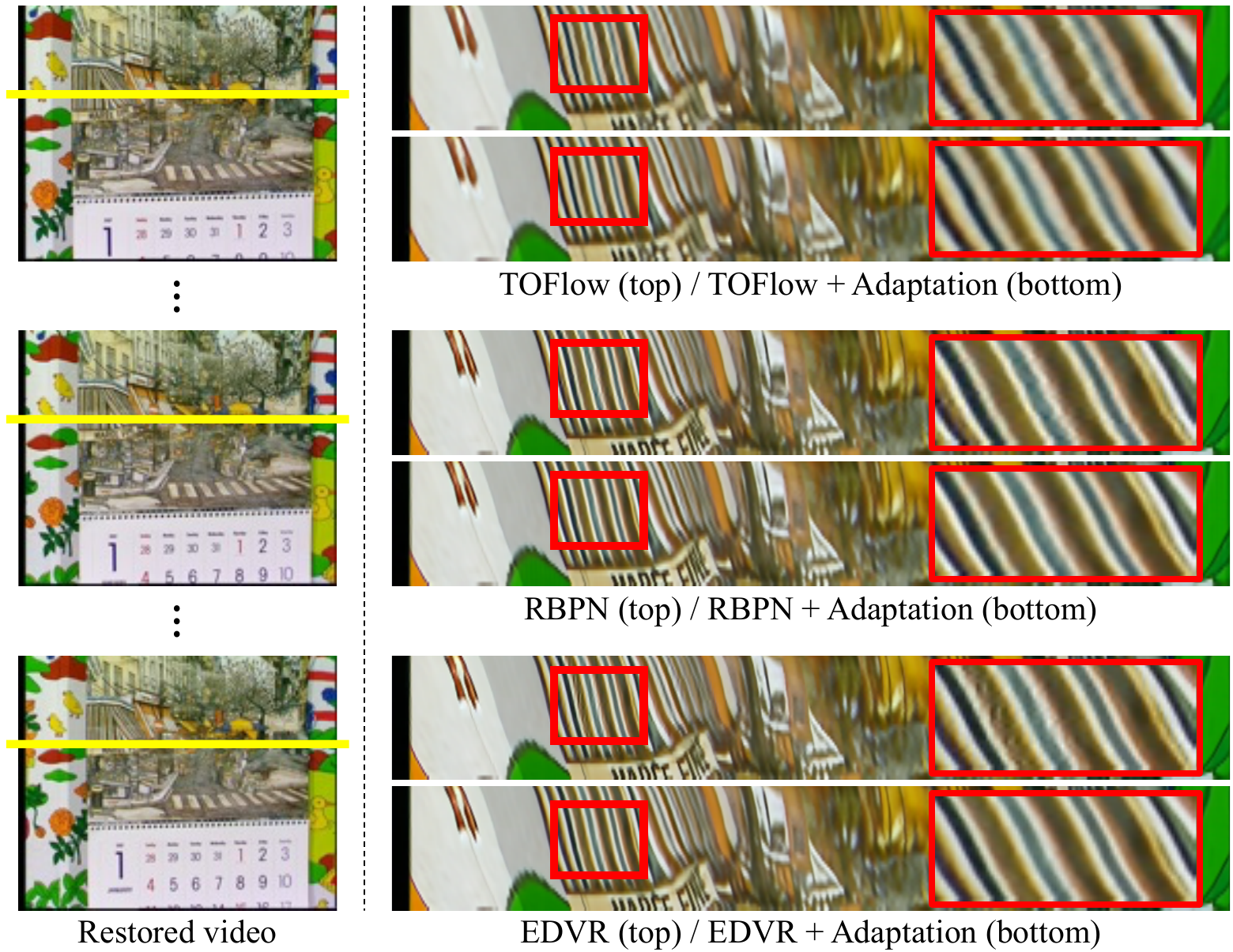}
\caption{Temporal profiles of the results from baselines and our adapted networks. Our proposed method improves the temporal consistency while preserving fine details.}
\label{fig_temporal}
\end{figure}

We also compare the temporal consistencies in \tabref{tab_tof} in terms of the correctness of estimated optical flow~\cite{chu2020tecogan}, and we see that our method consistently improves temporal consistency (\ie, tOF).
We also visualize the temporal consistency in \figref{fig_temporal}. 
We trace the fixed horizontal lines (yellow line in the left sub-figures) and vertically stack it for every time step.
Then, the noisy effect (\eg, jagged line) in the result indicates the flickering of the video~\cite{sajjadi2018frvsr}.
Thus, we conclude that the adapted networks achieve temporally more smooth results while maintaining sharp details over the baselines.

\paragraph{Efficient adaptation via knowledge distillation.}

\begin{table}[]
\centering
\footnotesize
\begin{tabular}{lccc}
\hline
\multirow{2}{*}{\textbf{Method}} & \multicolumn{2}{c}{\textbf{Adaptation cost}} & \multirow{2}{*}{\textbf{\begin{tabular}[c]{@{}c@{}}Performance\\ (PSNR/SSIM)\end{tabular}}} \\ \cline{2-3}
 & \textbf{GPU usage} & \textbf{Time/clip} &  \\ \hline
EDVR$_L$ & - & - & 27.35/0.8264 \\
EDVR$_S$ & - & - & 26.79/0.8087 \\ \hline
EDVR$_{L \rightarrow L}$ & 5.5GB & $\approx$10min & 27.53/0.8305 \\ \hline
EDVR$_{S \rightarrow S}$ & \multirow{2}{*}{3.2GB} & \multirow{2}{*}{$\approx$5min} & 27.05/0.8147 \\
EDVR$_{L \rightarrow S}$ &  &  & \textbf{27.41/0.8277} \\ \hline
\end{tabular}
\caption{Knowledge distillation results on the Vid4 dataset. Performance of the small student network is improved by leveraging pseudo datasets from the large teacher network. Left and right sides of the arrow indicate teacher and student, respectively.}
\label{tab_kd}
\end{table}

As conventional VSR networks are very huge, it takes much time to apply our adaptation algorithm at test time.
Thus, we reduce the adaptation time by using the knowledge distillation method in \algoref{algo_fa}.
We demonstrate the effects of our algorithm using the large and small versions of EDVR (EDVR$_L$ and EDVR$_S$ for each) on the Vid4 dataset. Note that, EDVR$_L$ has approximately 21 million parameters, and EDVR$_S$ has about 3.3 million parameters.

In \tabref{tab_kd}, we observe that we can reduce the adaptation time in half with less hardware resources by distilling knowledge from EDVR$_L$ to EDVR$_S$ (EDVR$_{L \rightarrow S}$) compared with the adaptation from EDVR$_L$ to EDVR$_L$ (EDVR$_{L \rightarrow L}$) while improving the performance over the large baseline network (EDVR$_L$).


This promising result opens an interesting research direction of combining knowledge distillation with the self-supervision-based SR task.

\subsection{Ablation study}
\label{4_3}

\paragraph{VSR quality and the number of recurring patches.}
\begin{figure}[]
\centering
\includegraphics[width=\linewidth]{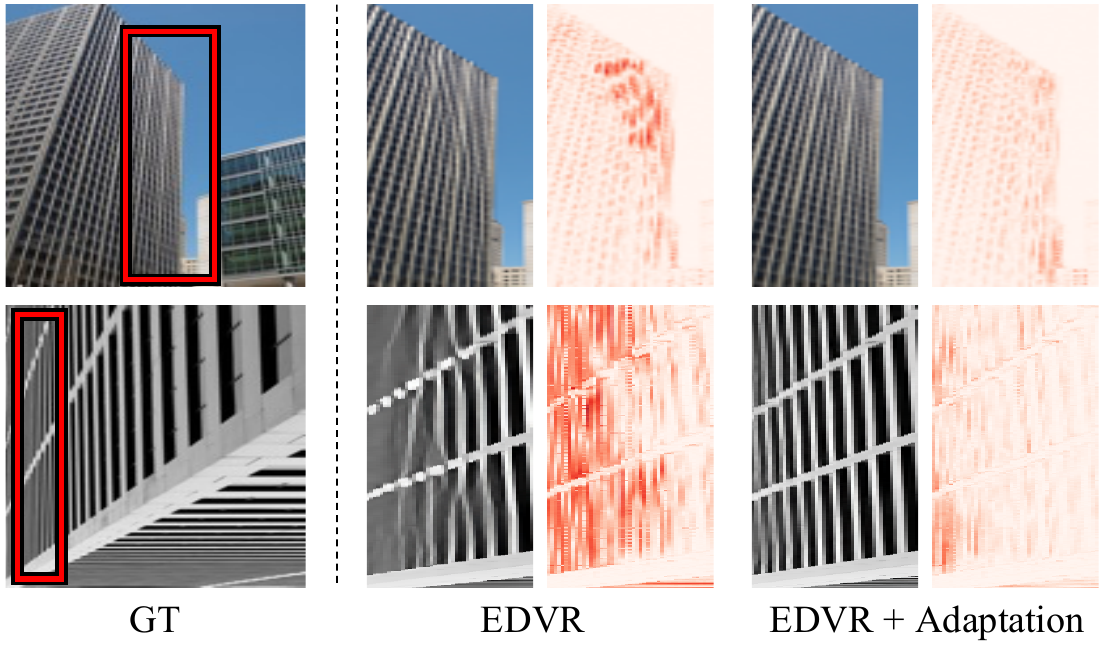}
\caption{SR results and absolute error maps. More redish color in the error map indicates higher error. Small structures are well restored with the aid of our adaptation while preserving the quality of large ones. See the difference between error maps before and after adaptation.}
\label{fig_map}
\end{figure}

To observe the enhanced region with our adaptation, we restore frames including highly repeated patches in \figref{fig_map}.
The error maps show that the smaller patches are well restored by our adaptation without distorting the larger patches; these results are in consistent with \thmref{thm1}. 

\begin{table}[]
\centering
\footnotesize
\begin{tabular}{lccc}
\hline
\multirow{3}{*}{\textbf{Dataset}} & \multirow{3}{*}{\textbf{EDVR~\cite{wang2019edvr}}} & \multicolumn{2}{c}{\textbf{EDVR~\cite{wang2019edvr} + Adaptation}} \\ \cline{3-4} 
 &  & \multicolumn{2}{c}{patch-recurrence} \\ \cline{3-4} 
 &  & low & \textbf{high} \\ \hline
Vid4 & 27.35/0.8264 & 27.40/0.8276 & \textbf{27.53/0.8305} \\
REDS4 & 32.45/0.8937 & 32.38/0.8927 & \textbf{32.69/0.8957} \\ \hline
\end{tabular}
\caption{Patch-recurrence and VSR results. Higher patch-recurrence produces better results on the Vid4 and REDS4 datasets according to \thmref{thm1}.}
\label{tab_algorithm}
\end{table}

Moreover, we measure the adaptation performance by changing the number of recurring patches.
For the comparison,
we first restore $T$ frames in the given video with $T$ different parameters, which are adapted for each frame without using neighboring frames (low patch-recurrence).
Next, we predicted results with a global parameter adapted using every frame in the given input video (high patch-recurrence).
These results are compared in \tabref{tab_algorithm}, and we see that we achieve better performance when the number of recurring patches is large. 
Notably, low patch-recurrence on the REDS4 dataset even degrades the performance over the baseline.

\paragraph{Random downscaling and VSR results.}

\begingroup
\setlength{\tabcolsep}{3.5pt}
\begin{table}[]
\centering
\footnotesize
\begin{tabular}{lcccc}
\hline
\multirow{4}{*}{\textbf{Dataset}} & \multicolumn{4}{c}{\textbf{Scaling for pseudo target generation}} \\ \cline{2-5} 
 & no downscale & \multicolumn{2}{c}{downscale} & upscale \\ \cline{2-5} 
 & \multirow{2}{*}{-} & fixed & \textbf{random} & random \\ \cline{3-5} 
 &  & 0.95 & \textbf{0.8} to \textbf{0.95} & 1.05 to 1.2 \\ \hline
Vid4 & \footnotesize{27.42/0.8280} & \footnotesize{27.48/0.8300} & \footnotesize{\textbf{27.53/0.8305}} & \footnotesize{27.35/0.8230} \\
REDS4 & \footnotesize{32.51/0.8931} & \footnotesize{32.61/0.8948} & \footnotesize{\textbf{32.69/0.8957}} & \footnotesize{32.31/0.8891} \\ \hline
\end{tabular}
\caption{VSR results by changing scaling methods for pseudo target generation with EDVR~\cite{wang2019edvr}. Downscaling in pseudo target generation improves the performance while upscaling degrades. Random downscaling shows the best performance.}
\label{tab_ds}
\end{table}
\endgroup

In \tabref{tab_ds}, we compare VSR results obtained with and without downscaling in generating the pseudo dataset.
We demonstrate that we can exploit self-similar patches by generating pseudo dataset with downscaling as illustrated in \figref{fig_corresponding}, and random downscaling records the best performance.
Note that no-downscaling and upscaling produce poor results since we cannot generate high-quality pseudo targets with more image details.

\paragraph{Application to SISR.}

\begin{table}[]
\centering
\footnotesize
\begin{tabular}{lcc}
\hline
\textbf{Dataset} & \textbf{\begin{tabular}[c]{@{}c@{}}RCAN~\cite{zhang2018rcan}\end{tabular}} & \textbf{\begin{tabular}[c]{@{}c@{}}RCAN~\cite{zhang2018rcan} + Adaptation\end{tabular}} \\ \hline
DIV2K & 30.77/0.8460 & \textbf{30.85/0.8473} \\
Urban100 & 26.82/0.8087 & \textbf{27.34/0.8198} \\ \hline
\end{tabular}
\caption{Quantitative results for SISR with RCAN~\cite{zhang2018rcan} using $\times$4 upscaling factor. Higher gain is acheived with higher patch-recurrence.}
\label{tab_sisr}
\end{table}

\begin{figure}[]
\centering
\includegraphics[width=\linewidth]{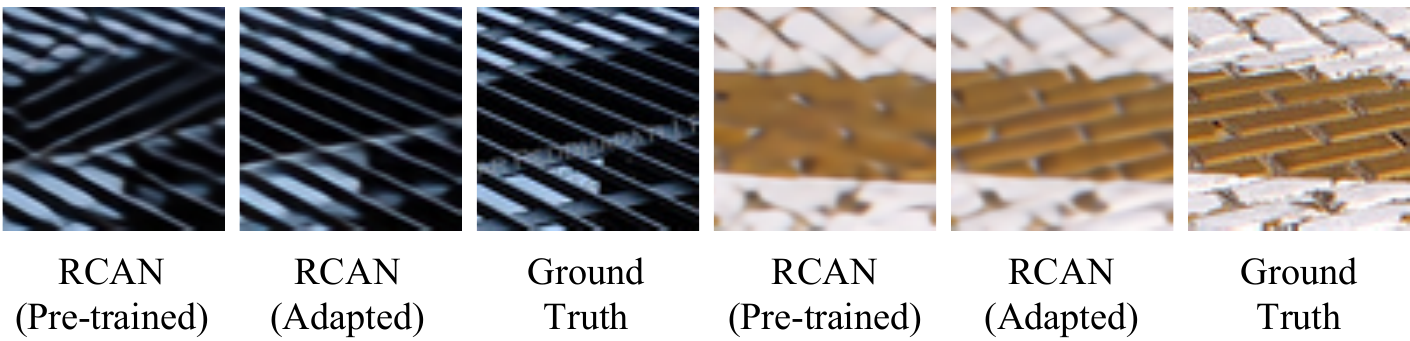}
\caption{Visual results for SISR with RCAN~\cite{zhang2018rcan}. Misaligned structures are well recovered with our adaptation.}
\label{fig_sisr}
\end{figure}

We conduct experiments to observe the applicability of our algorithm to the SISR task with RCAN~\cite{zhang2018rcan} on the DIV2K~\cite{agustsson2017div2k} and Urban100~\cite{huang2015selfex} datasets. 
Notably, RCAN is a state-of-the-art SISR approach currently.
We apply \algoref{algo_ssa} to RCAN by assuming that the the given video includes only a single frame.
\tabref{tab_sisr} shows the consistent improvements. 
In particular, on the Urban100 dataset, which contains highly recurring patches across different scales, performance gain is significant (+0.5 dB). 
Visual comparisons are also provided in \figref{fig_sisr}, we see the correctly restored edges with our adaptation. 

\section{Conclusion}

In this study, we propose a self-supervision-based adaptation algorithm for the VSR task. Although many SISR methods benefit from self-supervision, only a few studies have been attempted for the VSR task. Thus, we present a new self-supervised VSR algorithm which can further improve the pre-trained networks and allows to deal with large scaling factors by combining the information from the external and internal dataset. We also introduce test-time knowledge distillation algorithm for the self-supervised SR task. In the experiments, we show the superiority of the proposed method over various baseline VSR networks. 

\clearpage
{\small
\bibliographystyle{ieee_fullname}
\bibliography{egbib}
}

\end{document}